\documentclass[lettersize,journal]{IEEEtran}
\usepackage{amsmath,amsfonts}
\usepackage{algorithmic}
\usepackage{algorithm}
\usepackage{array}
\usepackage[caption=false,font=normalsize,labelfont=sf,textfont=sf]{subfig}
\usepackage{textcomp}
\usepackage{stfloats}
\usepackage{url}
\usepackage{verbatim}
\usepackage{graphicx}
\usepackage{booktabs}
\usepackage{multirow}
\usepackage{amsmath}
\usepackage{amssymb}
\usepackage{cite}
\DeclareMathOperator*{\argmin}{argmin}

\begin{document}

\title{Rotational Symmetry based Object Pose Estimation from Point Clouds in the Absence of Known 3D Models}

\author{
Weichen~Dai$^{1}$, 
Ruixun~Yu$^{1}$, 
Yangjie~Tang$^{1}$, 
Yifan~Du$^{1}$, 
Yiyang~Zhang$^{1}$, 
Donglei~Sun$^{2}$, \IEEEmembership{Member IEEE}
and Hua~Zhang$^{1*}$
\thanks{    
$^{1}$Key Laboratory of Brain Machine Collaborative Intelligence of Zhejiang Province, School of Computer Science, Hangzhou Dianzi University, Hangzhou 310018, China (zhangh@hdu.edu.cn)
$^{2}$Advanced Intelligent Manufacturing Research Group, the University of Nottingham Ningbo China, Ningbo 315100, China
*Corresponding author.
}
\thanks{Manuscript received April 19, 2021; revised August 16, 2021.}}

\markboth{Journal of \LaTeX\ Class Files,~Vol.~14, No.~8, August~2021}%
{Shell \MakeLowercase{\textit{et al.}}: A Sample Article Using IEEEtran.cls for IEEE Journals}


\maketitle

\begin{abstract}
Object pose estimation is crucial to many industrial applications, with one example being the automated spray painting with a robot. However, confidentiality concerns often limit access to high-quality 3D models, posing a significant challenge for pose estimation based on point clouds. In such scenarios, rotational symmetry—a readily accessible characteristic of many industrial objects—can provide valuable prior information to facilitate pose estimation. 
In this paper, a method is proposed to leverage the rotational symmetry commonly found in industrial objects to address the challenge caused by the absence of 3D models. 
The pose of the object is jointly estimated with point cloud refinement in iterations.  This iterative optimization relies on the loss of the constraint of rotational symmetry. To construct such loss of rotational symmetry, each 3D point is rotated based on the currently estimated pose, and multiple correspondences are then identified using the nearest neighbor search through exploiting the property of rotational symmetry. 
The identified correspondences are used to compute the rotational symmetry constraint loss, which iteratively refines both the pose and the point cloud. By explicitly incorporating rotational symmetry into the optimization process, the proposed method achieves robust pose estimation and generalizes well across diverse object types.
The proposed method is evaluated using data specifically created from point clouds without known 3D models, which comprises a dataset from four types of synthetic objects and one real wheel hub from the production line. 
Experimental results demonstrate that the proposed method achieves a performance comparable to methods relying on known 3D models.
\end{abstract}


\section{Introduction}
\IEEEPARstart{O}{bject} pose estimation calculates both the rotation and translation of an object with respect to a world-fixed coordinate frame~\cite{wu2024recurrent}. Accurate calculation of object poses is a crucial prerequisite for various downstream applications, such as autonomous robotic spray painting~\cite{gleeson2022generating} and manipulation~\cite{zhang2021practical}. Recently, considerable attention has been directed towards the methods of determining object poses from point clouds~\cite{hoang2022voting,drost2010model,gao20206d,guo2014integrated}, which can be generated from various sources, such as laser range finders, industrial high-resolution 3D sensors, and multi-view stereo cameras, to name but a few. 
Accurate pose estimation from 3D point clouds is still challenging due to various factors such as measurement noise, incomplete sampling, and the presence of occlusions~\cite{xing2021efficient}.

With impressive progress made in deep methods, object pose estimation methods have developed rapidly and achieved good performance~\cite{xiang2017posecnn,wang2019densefusion}. Popular methods of pose estimation from point clouds are built upon finding the correspondence using global or local feature matching between the input data and a model of the object~\cite{hinterstoisser2013model,wang20216d,he2020pvn3d} created from either computer-aided design~(CAD) or scanning. The pose is then estimated by using these identified correspondences. The estimation result can be further refined by using an Iterative Closest Point (ICP) algorithm if the 3D model of the object is available. However, high-quality 3D models of industrial objects are not commonly available, often due to confidentiality concerns, and hence the exploration of model-free methods on object pose estimation from point clouds is necessary.

\begin{figure}[t!]
    \centering
    \includegraphics[width=1.0\linewidth]{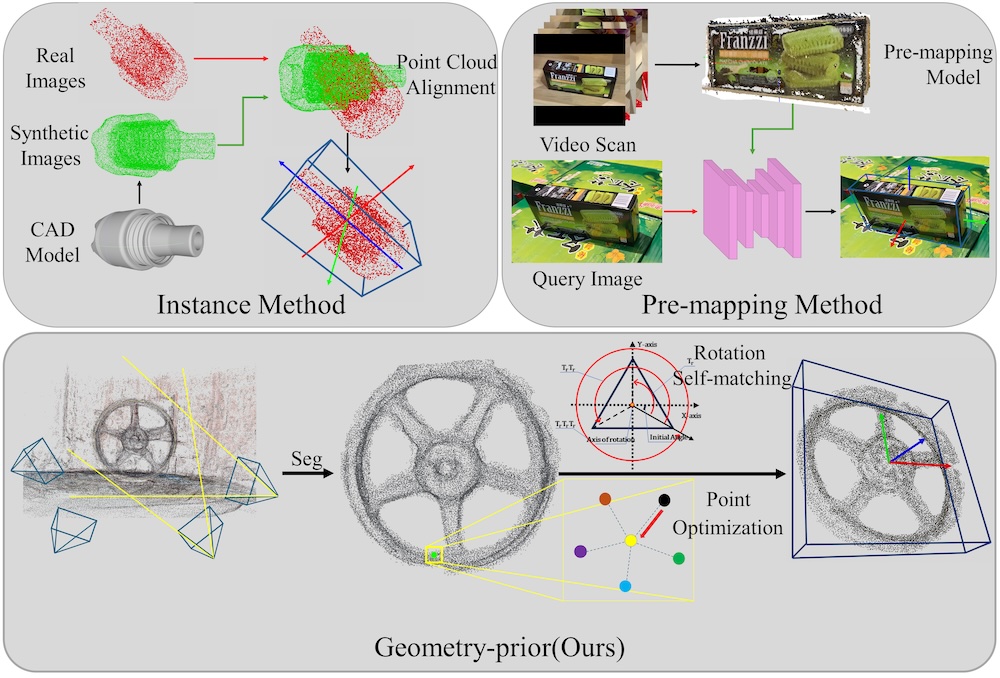}
    \caption{Comparison of object pose estimation methods with and without using known 3D models. To address the model availability challenge, 
    unlike previous pre-mapping methods that build the 3D model in advance, the proposed method utilizes rotational symmetry commonly found in industrial objects for object pose estimation.}
    \label{fig:headpic}
\end{figure}

To reduce the demand for high-quality 3D models and improve the generalization capability~\cite{sun2022onepose}, this study pays more attention to the utilization of inherent geometric characteristics, which can offer valuable prior model knowledge for object pose estimation. Similar to the problem setting without known models~\cite{li2022ws}, simultaneous localization and mapping~(SLAM)~\cite{cadena2016past} from 3D sensors assumes that only a sequence of 3D measurements is available in an unknown environment, and a point cloud model can be reconstructed based on the geometric consistency between the multi-view measurements. Then, the poses of sensors can be estimated within the reconstructed point cloud model. Similar to SLAM, the geometric characteristics of objects also provide prior model information, and the raw point clouds can be further refined from multiple perspectives, through which the object pose can be estimated. 

Given the specific operational characteristics and design considerations inherent in the processing of industrial objects~\cite{siemiatkowski2007modelling}, for example, the rotational machining process carried out by a lathe, a finding emerges: many industrial objects possess a discernible degree of symmetry, especially rotational symmetry. Though this symmetry may have some negative impact as it introduces the ambiguity issue~\cite{pitteri2019object,duffhauss2023symfm6d}, it provides outweighing benefits: not only can it offer invaluable prior information to facilitate object pose estimation but it also serves as a more readily accessible source of information, considering the limited availability and inconsistent quality of CAD models.

In this paper, we propose a object pose estimation method from point clouds which exploits the rotational symmetry of industrial objects to address the problem of unavailability of 3D models. Given the raw point cloud, the object point cloud is segmented from the surface where the object is positioned by point clustering of the surface. Then the pose and point cloud of the object are iteratively estimated, leveraging the constraint of rotational symmetry. Each 3D point is transformed to the object coordinates using the estimated pose. Afterwards, since the clutter and occlusion generate indiscrimination, multiple correspondences will be identified using the nearest neighbor search after several rotations, exploiting the property of rotational symmetry that an object retains its shape after a rotation. These correspondences are used to refine the pose and point cloud in each iteration. This conventional pipeline leverages the explicit rotational symmetry of the object, enabling the utilization on the fly for an unseen instance and enhancing generalization capabilities across diverse objects. To evaluate the proposed method, a dataset is collected from objects with rotational symmetry for pose estimation from point clouds,  including point clouds of four synthetic objects and a real wheel hub.

The main contributions of this work are as follows.
\begin{itemize}
\item A method is proposed that can utilize rotational symmetry to achieve object pose estimation without 3D models.
\item A method for joint point cloud refinement and object pose estimation is developed that can further improve the accuracy of object pose estimation.
\item The proposed method shows competitive performance on the dataset, and shows a better generalization capability to arbitrary objects thanks to the explicit geometric constraint.
\end{itemize}







\section{Related Work}

\begin{figure}
    \centering
    \includegraphics[width=0.9\linewidth]{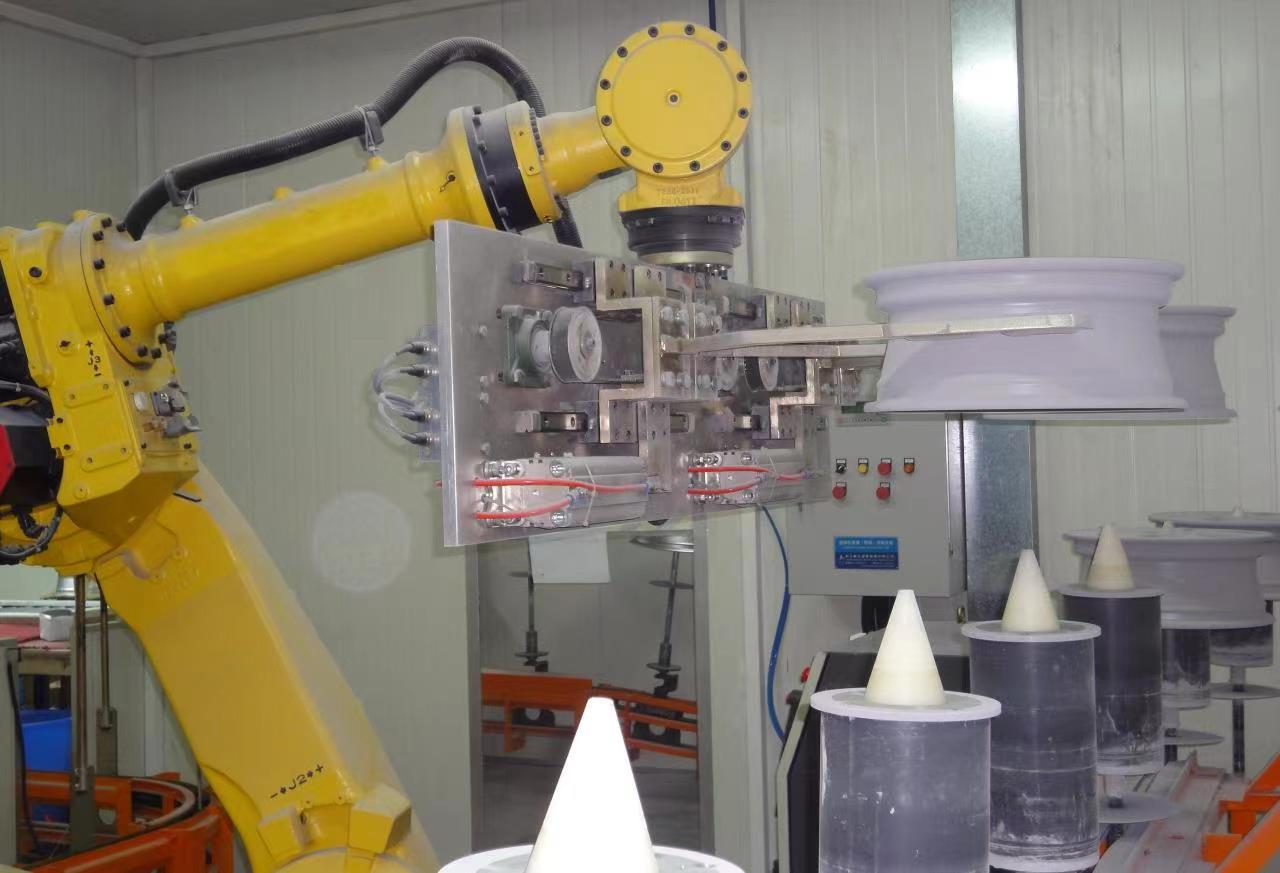}
    \caption{ Spraying operation from the factory. The objects are fixed with clamps.}
    \label{fig:factory}
\end{figure}

Object pose estimation~\cite{xiang2017posecnn,he2019sparse,zhang2022eanet,cheng2021real} has wide applications in various industrial processes~\cite{zhang2021practical}. For instance, in industrial production and manufacturing, the spraying operation of parts plays a crucial role~\cite{gleeson2022generating}. In traditional automated production processes, the preset pose is obtained by fixing the objects with limiting accessories such as clamps, which allows for the automated production to proceed and ensures economic benefits, particularly in mass production scenarios, as shown in Fig.~\ref{fig:factory} taken in the factory.
However, when processing non-standard or small batches of objects, the repeated design, manufacturing and assembly of new accessories increase the production costs. In addition to sensing the object shape from the point clouds of sensors, accurate object pose estimation will address the challenge of the aforementioned high costs and optimize the operational tasks for such scenarios.

\subsection{Object Pose Estimation from Point Clouds}



Most conventional methods of object pose estimation  primarily rely on feature-based matching, including both local and global features. The methods based on global features usually describe the entire point cloud of the object with self-similar surface parts, such as planar patches, as a feature~\cite{aldoma2011cad,rusu2010fast, wohlkinger2011ensemble}.
The methods based on local features mainly explore the local structural information around some key points on the objects~\cite{guo20143d}. Therefore, object detection can be achieved in the cluttered scenes. There are some methods that combine the advantages of both types of features~\cite{drost2010model} and a Point Pair Feature (PPF) has been proposed to describe the relative position and orientation of two oriented points. Following this promising idea, many works~\cite{guo2021efficient,xing2021efficient} focus on the improvment of various aspects, including pipline~\cite{Birdal2015pipeline}, sampling strategy~\cite{hinterstoisser2016going}. In recent years, methods utilizing the capabilities of deep learning have been employed to achieve object pose estimation~\cite{gao20206d,zhou20226}. However, these methods often rely on a known 3D model, and due to confidentiality reasons, designers or manufacturers are usually unwilling to provide such models, thus making the implementation of these methods in downstream industry difficult if not impossible.

\subsection{Object Pose Estimation without Known 3D Models}

Compared to instance-level pose estimation, which needs a known 3D object model in implementation, many recent methods work on category-level object pose estimation~\cite{ahmadyan2021objectron,wang2019normalized,yu2024synthetic} to meet the demands in practical applications. Most of these methods train a network on various instances of the category. Therefore, the network can learn a category-level representation of object appearance and shape and thus is able to generalize to new instances in the same category. In addition to category-level object pose estimation, pre-mapping methods try to first construct a model using sensing information, as shown in Fig.~\ref{fig:headpic}, and then perform pose estimation based on the constructed model, see, for example, the Structure from Motion (SfM)~\cite{sun2022onepose} and comparison networks~\cite{liu2022gen6d}. However, such methods still require three-dimensional models or data with known poses during the training or construction step and do not leverage readily available prior geometry information inherent in the symmetry of the objects.

\subsection{Symmetry-Aware Object Pose Estimation}


\begin{figure}[t!]
    \centering
    \includegraphics[width=\linewidth]{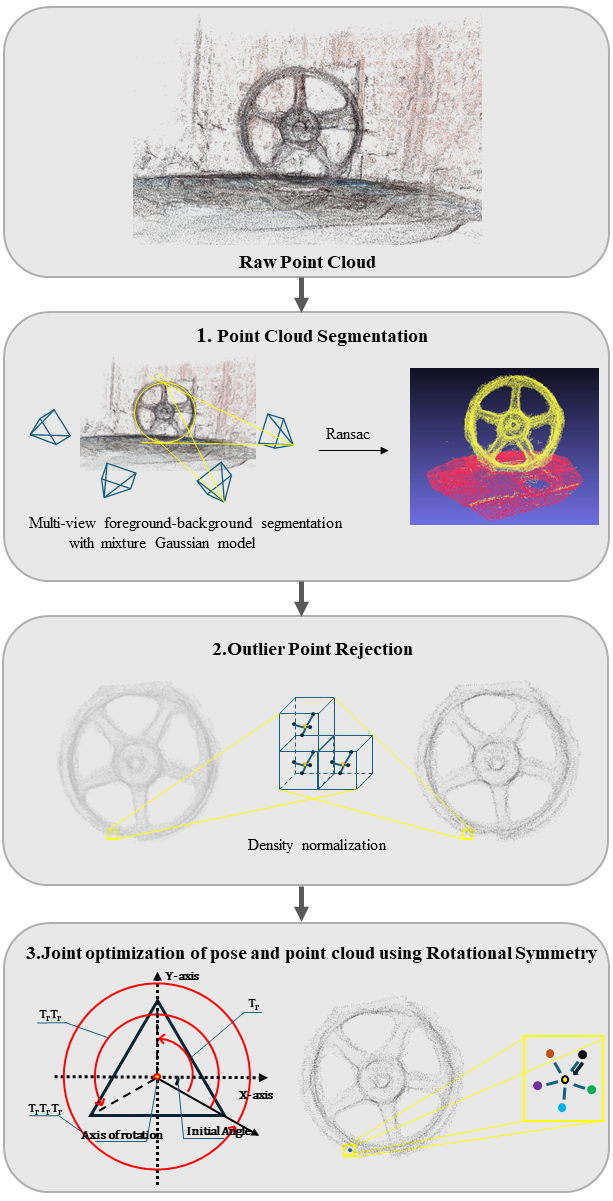}
    \caption{ System overview. }
    \label{fig:overview}
\end{figure}

Due to the ambiguity of symmetric objects, multiple poses can be inferred with a specific appearance for them. To address this issue, an Average Distance for Symmetry (ADD-S) metric or a symmetry-invariant pose distance metric is adopted to train the networks~\cite{xiang2017posecnn,wang2019densefusion,mo2022es6d}. Other methods propose an additional output to classify or predict the symmetry~\cite{pitteri2019object,rad2017bb8,zhang2020symmetry}. In addition, symmetry-aware key-point detection~\cite{duffhauss2023symfm6d} and symmetry-aware matching loss~\cite{zhao2023learning} were proposed to further improve the performance. However, these methods primarily focus on minimizing the impact of rotational symmetry on pose estimation when known 3D models are available. Hence in-depth exploration of prior information from rotational symmetry to enhance or achieve object pose estimation in scenarios with unknown geometric structures remains valuable.

\section{Method}

In the industrial scenario of robotic painting, spraying is performed on one object at a time. Therefore, without deep networks, a multi-view point cloud segmentation method is employed to divide the object point clouds. Subsequently, based on the segmentation results, a pose estimation will be conducted for the object.
The proposed method consists of steps for point cloud segmentation, outlier removal, and joint optimization of point clouds and object poses, as depicted in Fig.~\ref{fig:overview}.

\subsection{Point Cloud Segmentation}

This section introduces the multi-view foreground-background segmentation with a mixture Gaussian model and a bottom-surface removal component, which identifies and excludes the background and bottom-surface points present in the raw point cloud to only keep the point cloud belonging to the object.

\subsubsection{Multi-view foreground-background segmentation with a mixture Gaussian model} 
\begin{figure}[t!]
    \centering
    \includegraphics[width=0.36\linewidth]{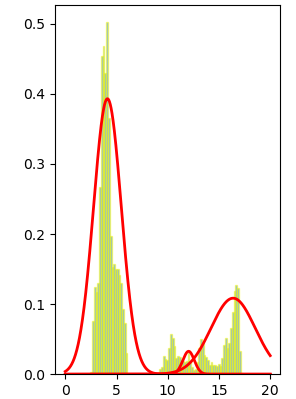}
    \includegraphics[width=0.58\linewidth]{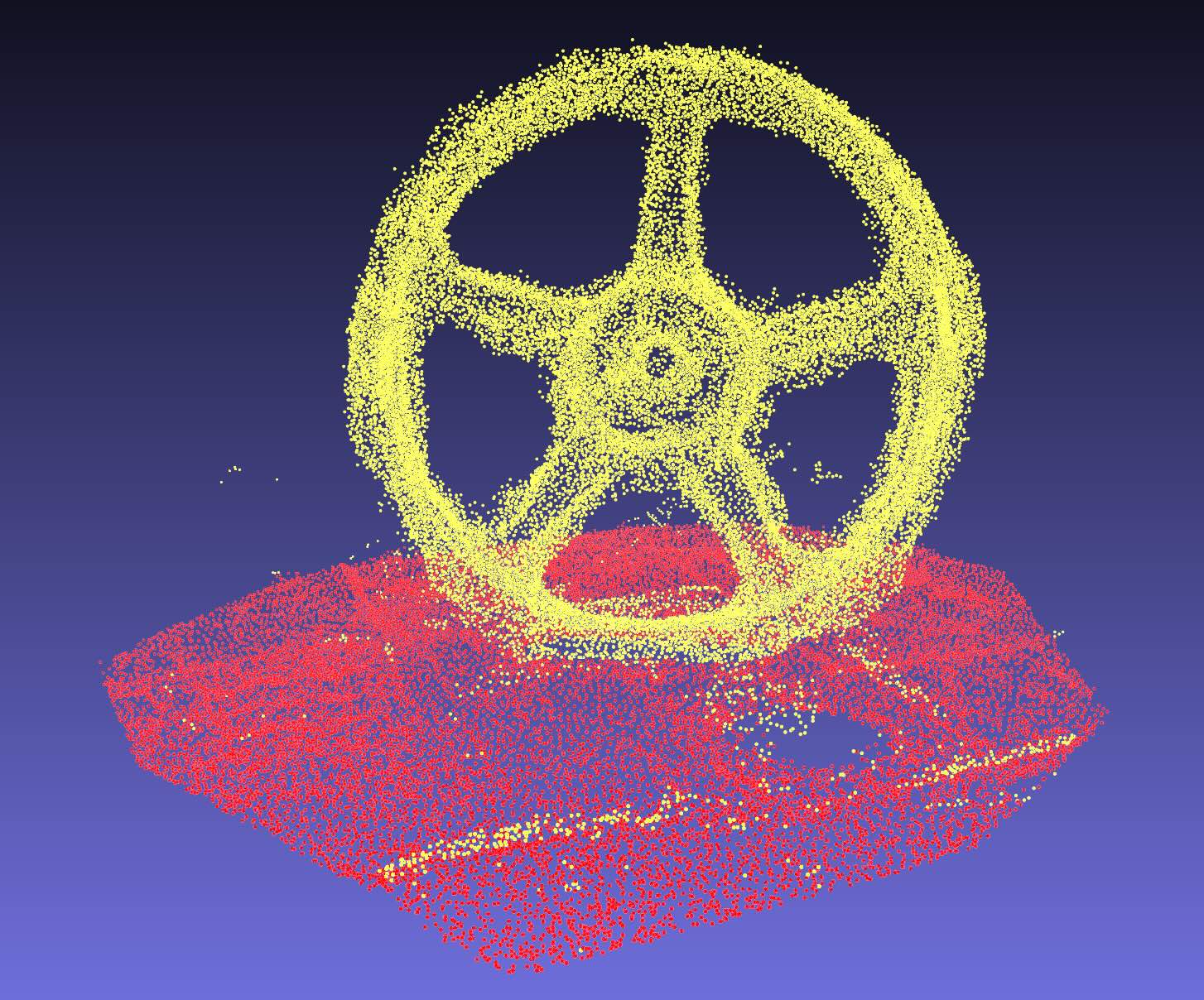}
    \caption{ Left: Gaussian clustering on histogram of depth statistics. Right: Segmentation results using surface-plane fitting.}
    \label{fig:seg}
\end{figure}

The raw point cloud, denoted by a point set $P$, can be obtained by depth measurements or SfM. Since the originally collected point clouds often capture not only the area where the object is located but also its background information, to remove the background point cloud $P^{bg} \subset P$, a multi-view foreground-background segmentation method is proposed. All points from the point cloud are projected onto a specific view
\begin{equation}
    \mathbf{u}_i,d_i = \pi(\mathbf{T}_{cw}\mathbf{p^w_i})
\end{equation}
where $\mathbf{T}_{cw}$ is the transformation matrix that converts the $i$-th point from the world coordinates $\mathbf{p^w}$ to the camera coordinates.
$\pi \colon \mathbb{R}^3 \mapsto \mathbb{R}^3$ represents the camera model with intrinsic parameters which performs a reprojection of the $i$-th 3D point onto the frame in terms of the image coordinates $ \mathbf{u}_i \in \mathbb{R}^2$ with depth $d_i$.
Then, all depth results are calculated and saved as a histogram of depth value for the point cloud in that view. 
Since there is a significant depth difference between the foreground and background points, as illustrated in Fig.~\ref{fig:seg}, the proposed method creates clusters of point clouds in each view based on the depth statistics using a Gaussian mixture model
\begin{equation}
    GM = \sum_{i=1}^{K}\phi_{i}{\mathcal {N}}({\boldsymbol {\mu_{i},\Sigma_{i}}})
\end{equation}
where the $i$-th vector component is characterized by a normal distribution with weight ${ \phi _{i}}$, mean value ${ {\boldsymbol {\mu _{i}}}}$ and covariance matrix ${ {\boldsymbol {\Sigma _{i}}}}$.
Point cloud clusters with a smaller mean in the Gaussian statistics are considered the foreground, thereby creating masks for individual views. Each point $\mathbf{P}_i$ outside the mask will be eliminated from the point set $P$. Subsequently, multiple views will undergo the clustering segmentation process multiple times to ultimately obtain the point cloud where the object is located.

\subsubsection{Bottom surface removal}

In an actual scene, the object is typically placed on a plane, resulting in the ground point clouds and object point clouds being closely intermingled. Without any prior information, the segmentation will be challenging.
Assuming that the ground forms a large flat surface which exists on the object but with less significance than the ground.
Due to potential noise in the point cloud, the Hough space may not be suitable. Therefore, a plane-fitting method using RANSAC is utilized to identify the ground plane from the point cloud above. 
The plane parameters are optimized based on the sum of point-to-plane distances
\begin{equation}
\mathbf{v} \leftarrow \argmin_{\mathbf{v}}
\sum_{i}\left\Vert \frac{\left| \begin{bmatrix}a & b & c &d \end{bmatrix} \mathbf{P}_i^w \right|}{\sqrt{a^2+b^2+c^2}}\right \Vert
\end{equation}
where $ \mathbf{v}^{\top} = \begin{bmatrix}a & b & c &d \end{bmatrix}$ denotes the parameters of the surface plane $ax+by+cz+d=0$. Then, the RANSAC iterations are performed to find the inlier points of the plane model.
Subsequently, all point clouds representing this ground are removed to obtain the required initial object point cloud.

\subsection{Outlier Point Rejection}

This section presents the outlier-rejection method which enhances the quality of the object point cloud by eliminating outlier points.

\subsubsection{Density normalization} 

Due to factors such as varying distances of measurement, on-site lighting conditions, and the surface material of the object, the density of the constructed object point cloud is often uneven. Additionally, to further enhance accuracy, the point cloud is downsampled using grid normalization. This involves partitioning the space into grids and averaging the positions of all the points inside each as the normalized result for that grid.

\subsubsection{Point cloud filtering} 

The downsampled point cloud effectively reduces noise, and yet residual outlier points persist. These anomalies stem from the specular reflections incorporated during point cloud reconstruction and the remnants from the uneven ground-surface removal.
To address this issue, both the Region of Relevance (ROR) filter and Statistical Outlier Removal (SOR) filter are sequentially applied to the point cloud. This combination retains the object point cloud while effectively eliminating the outlier points.



\subsection{Joint Optimization of Pose and Point Cloud using Rotational Symmetry}





The axis of symmetry in objects with rotational symmetry is defined as the pose direction and the center of mass serves as the pose position.
Based on the point cloud of the object provided by the aforementioned processes, the initial orientation and position of the object pose $\hat{\mathbf{T}}$ are determined respectively by fitting a plane and calculating the centroid of all points.


\subsubsection{Joint optimization of point cloud and pose}

Due to the rotational symmetry of the object, any point on the object should still remain on the object after a rotation around the axis of symmetry.
For objects that are discretely rotationally symmetric of order $n$, the rotation transformations that constitute rotational symmetry form a set
\begin{equation}
    {T}_r=
    \{
    \mathbf{T}_r, \mathbf{T}_r \mathbf{T}_r, \mathbf{T}_r \mathbf{T}_r \mathbf{T}_r,\dots,\underbrace{\mathbf{T}_r \mathbf{T}_r \cdots\mathbf{T}_r}_{n-1}
    \}
    \label{T_}
\end{equation}
where $\mathbf{T}_r$ denotes the rotation transformation of $(360^\circ/n)$ 
around the axis of symmetry.
With the initial pose $ \mathbf{T} \leftarrow \hat{\mathbf{T}}$, similar to the Iterative Closest Point (ICP) algorithm that aligns two point clouds, the object point cloud is rotated to find the corresponding points, thereby estimating the object pose $\mathbf{T}$ through the self-alignment of the object point cloud under multiple symmetric rotations
\begin{equation}
{\mathbf{T}},\mathbf{p}_i^w = \argmin_{\mathbf{T},\mathbf{p}_i^w}
\sum_{i} \sum_{\mathbf{T}' \in T_r} 
\left\Vert 
\mathbf{p}'_i -  \mathbf{T}' \mathbf{T} \mathbf{p}_i^w
\right \Vert_\gamma
\end{equation}
where $\mathbf{T} \mathbf{p}_i^w $ transforms the $i$-th point to the object coordinates. $\mathbf{T}'$ denotes a rotation transformation in the set ${T}_r$. $\mathbf{p}'_i$ denotes the closest point in the original object coordinates to $\mathbf{T}' \mathbf{T} \mathbf{p}_i^w$. The Huber loss $\gamma$ is applied to improve the robustness of the estimation. It should be noted that the object pose and the point cloud are optimized iteratively. The proposed loss traverses all points on the object point cloud and calculates the distance to their nearest point after the rotation. Therefore, each point has $n$ observations, which increases the utilization of all frames. If the estimations of the object pose and points are accurate, the loss will be minimized. After the optimization, the original object point cloud is augmented by multiple rotations to fill in the missing parts caused by occlusion. Then, the translation $\textbf{t}$ of the matrix $\mathbf{T}=\begin{bmatrix}
    \mathbf{R} & \mathbf{t} \\
    \mathbf{0} & 1
\end{bmatrix}$ is updated with the new centroid of the augmented object point cloud, where $\mathbf{R}$ is the rotation matrix.

\subsubsection{Estimation of primary orientation}

\begin{figure}[t!]
    \centering
    \includegraphics[width=\linewidth]{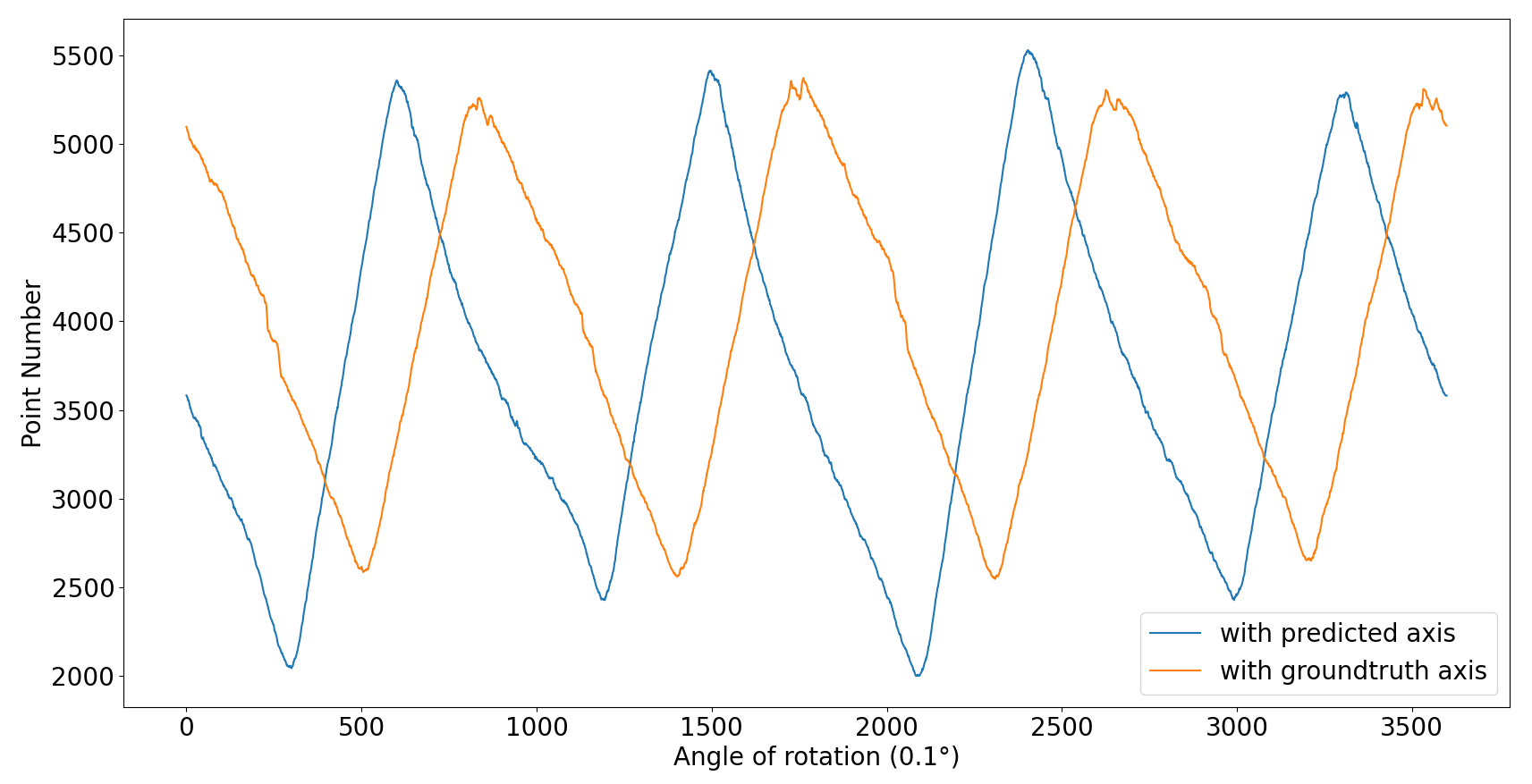}
    \caption{Primary-orientation estimation with extreme points. With the axis of rotational symmetry, the same yaw angles of the ground-truth point cloud and the optimized point cloud can be aligned.}
    \label{fig:yaw}
\end{figure}

The estimated direction of the augmented point cloud is chosen as the primary orientation of the object. Since the estimation result does not contain yaw angle, the histogram of point clouds in different directions is counted based on the augmented point cloud. When the axis of rotational symmetry is known, statistical analysis of the optimized point cloud at different angles reveals a periodic pattern. Define the yaw angle of the ground-truth point cloud as the extreme point. Then, statistical analysis of the point cloud based on the initial estimation of the axis of rotational symmetry also yields locally aligned extreme points, as shown in Fig.~\ref{fig:yaw}. Therefore, the direction with the extreme value of the point cloud is selected as its primary orientation.

\subsubsection{Implementation}

Each point needs to be rotated multiple times to find its corresponding point and this creates a massive computational burden. Therefore, in implementation,  the loss function is adjusted and a KD tree is utilized to accelerate the search for the nearest point in the point cloud
\begin{equation}
\mathrm{loss} = KD(\mathbf{p}_i^w)-  \mathbf{T}^{-1} \mathbf{T}' \mathbf{T} \mathbf{p}_i^w
    \label{KDlose}
\end{equation}
where the $KD(\cdot)$ function can provide the corresponding point $ \mathbf{p}'^w_i = KD(\mathbf{p}_i^w) $. Since the KD tree can be constructed based on the original point cloud, this eliminates the need to recreate it during each iteration of the optimization when $ \mathbf{T}$ is updated, thus reducing the computational burden.


\section{Experiment}


\subsection{Dataset Acquisition}

Since there is currently no dataset specifically designed for object pose estimation from the point clouds of a single object, the constructed dataset for evaluation has two parts, a synthetic dataset and some wheel-hub data. The most significant difference between the two datasets is that the synthetic dataset includes the ground truth point cloud and its object pose. The centroid of the object's shape is chosen as the coordinate origin of the object pose and the axis of symmetry serves as the $z$-axis.

\subsubsection{Synthetic dataset }

\begin{figure}[t!]
    \centering
    \includegraphics[width=\linewidth]{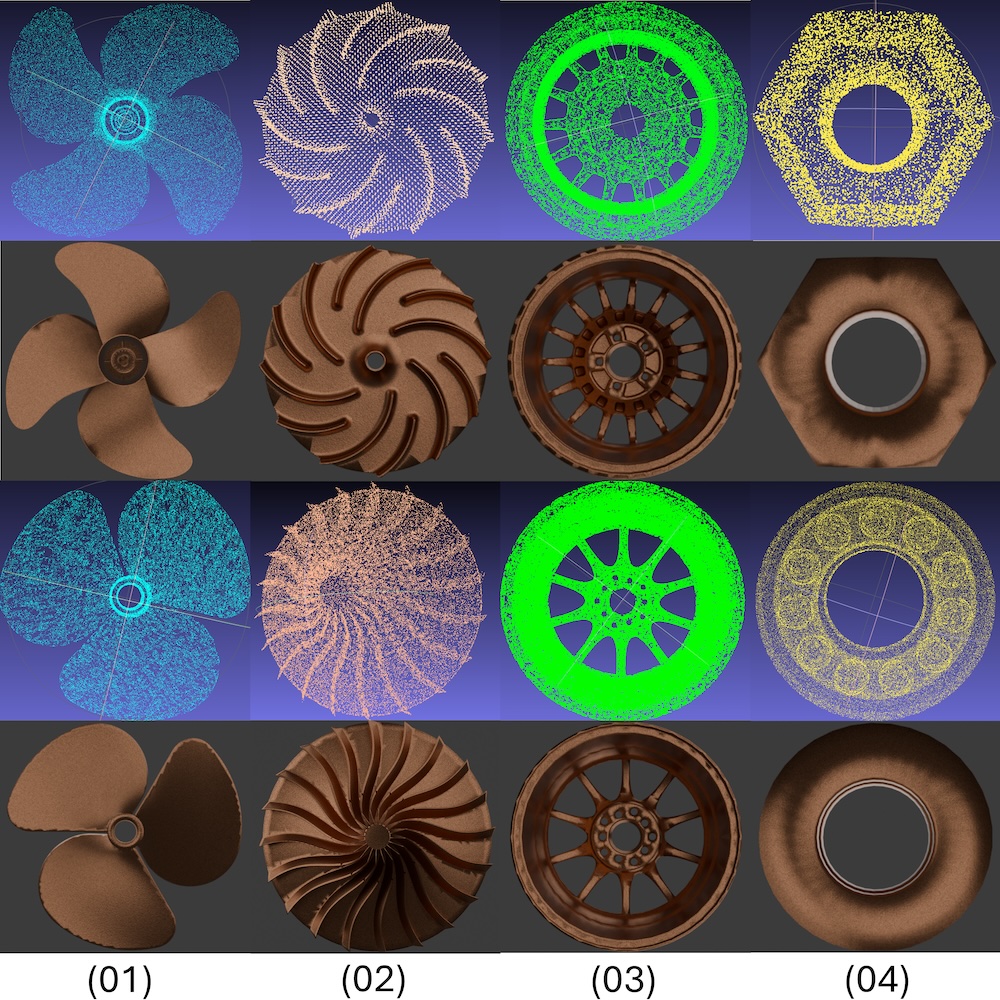}
    \caption{Synthetic Dataset. Objects (lower) are represented alongside their corresponding ground-truth point clouds (upper).}
    \label{fig:dataset}
\end{figure}

The data in the synthetic dataset was sampled based on CAD models. To diversify the dataset, CAD models of four different types of objects with rotational symmetry were used for data collection, including propellers, impellers, wheel hubs, and small parts (total 14 pcs), as shown in Fig.~\ref{fig:dataset}. To ensure that the accuracy of the estimated objects among different types was at the same scale and comparable, all these objects were normalized to have a diameter of 1. Using these CAD models, complete point clouds were initially obtained by random sampling as ground truth. Subsequently, with random views, noise was introduced to each captured color and depth image to simulate the noisy real-world scenarios with missing data due to occlusions. Additionally, to effectively evaluate the proposed method,  various noise levels were considered when creating the synthetic data. 

\subsubsection{Wheel-hub data}

\begin{figure}[t!]
    \centering
    \includegraphics[width=\linewidth]{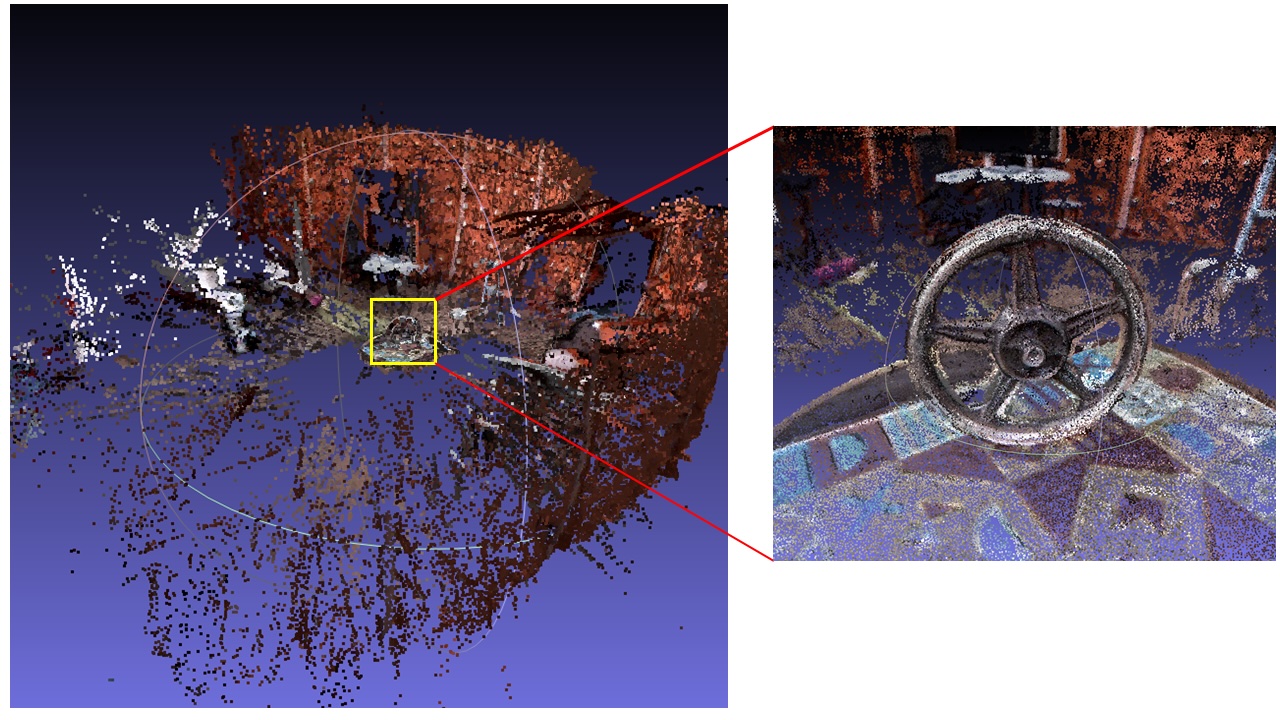}
    \caption{Point cloud of the wheel hub constructed with the SfM method.}
    \label{fig:wheel}
\end{figure}

This part of the raw data was collected from a real wheel hub from the factory. To better evaluate the performance of the proposed method, an SfM approach based solely on RGB data was employed for point cloud reconstruction, as shown in Fig.~\ref{fig:wheel}. In comparison to the point cloud data directly measured by depth cameras, point clouds obtained through visual SfM methods exhibit more nonlinear noise. This is particularly evident for wheel hubs as they are characterized by metal surfaces with minimal surface texture, resulting in a point cloud with increased error. It is important to note that the original SfM-generated point cloud contains both foreground and background points. The latter have been removed through point cloud segmentation but the former still remain in the cloud and contain substantial noise as well as the base point cloud where the wheel hub was positioned.

\subsection{Evaluation Metric}

Due to the presence of objects with rotational symmetry, the proposed method is evaluated using the metric of Average Distance for Symmetry. This metric computes the mean of the closest-point distances between the points transformed by the ground-truth pose and the estimated pose:
\begin{equation}
    \mathrm{ADD\text{-}S} = \frac{1}{p} \sum_{ \mathbf{x}_1 \in \mathcal{P}} \min_{\mathbf{x}_2\in \mathcal{P}} || (\mathbf{R}\mathbf{x}_1+\mathbf{t}) - (\mathbf{R}^{gt}\mathbf{x}_2+\mathbf{t}^{gt}) ||
\end{equation}
where $\mathbf{R}^{gt}$ and $\mathbf{t}^{gt}$ denote the ground-truth rotation and translation, respectively, $\mathbf{R}$ and $\mathbf{t}$ denote the estimated rotation and translation by the method under evaluation, $\mathcal{P}$ denotes the set of 3D points, and $p$ is the number of points.



\subsection{Quantitative Comparison}


\subsubsection{Accuracy}

In this experiment, the four types of objects in the synthetic dataset were individually computed for ADD-S and then averaged.
As shown in Table~\ref{table:comp_mwm}, the average ADD-S for the four types of objects 
proves that the proposed method (denoted by OURS) can provide comparable accuracy to the ICP method. Particularly noteworthy are the results for impellers, which exceed those of ICP, indicating that exploiting rotational symmetry as a prior information is effective. 
The unsatisfactory performance of Gen6D~\cite{liu2022gen6d} can be attributed to the inherent limitations of reconstructing objects with minimal textures, therefore it was struggling to extract information from the images of those objects. 

\begin{table}[t!]
    \centering
    \caption{Comparison of average ADD-S on the synthetic dataset. The best results are in bold.}
    \label{table:comp_mwm}
    \resizebox{\linewidth}{!}
{
    \begin{tabular}{ll|cccc}
    \toprule
         & Method  &Wheel hubs & Impellers & Parts  & Propellers     \\ \midrule
        \multirow{1}{*}{w/ 3D Models} 
        &\multirow{1}{*}{ICP}   & \textbf{0.015} & 0.032 & \textbf{0.007} & \textbf{0.031}\\ 
        
        \midrule
        \multirow{2}{*}{w/o 3D Models} 
        &\multirow{1}{*}{Gen6D}  & 5.253  & 12.160 & 29.164  &  3.341\\ 
        
        &\multirow{1}{*}{OURS}   & 0.031 & \textbf{0.017} & 0.013 & 0.036 \\ 
        \bottomrule
    \end{tabular}
    }
\end{table}







\subsubsection{Robustness to noise}

\begin{figure}[t!]
    \centering
    \includegraphics[width=\linewidth]{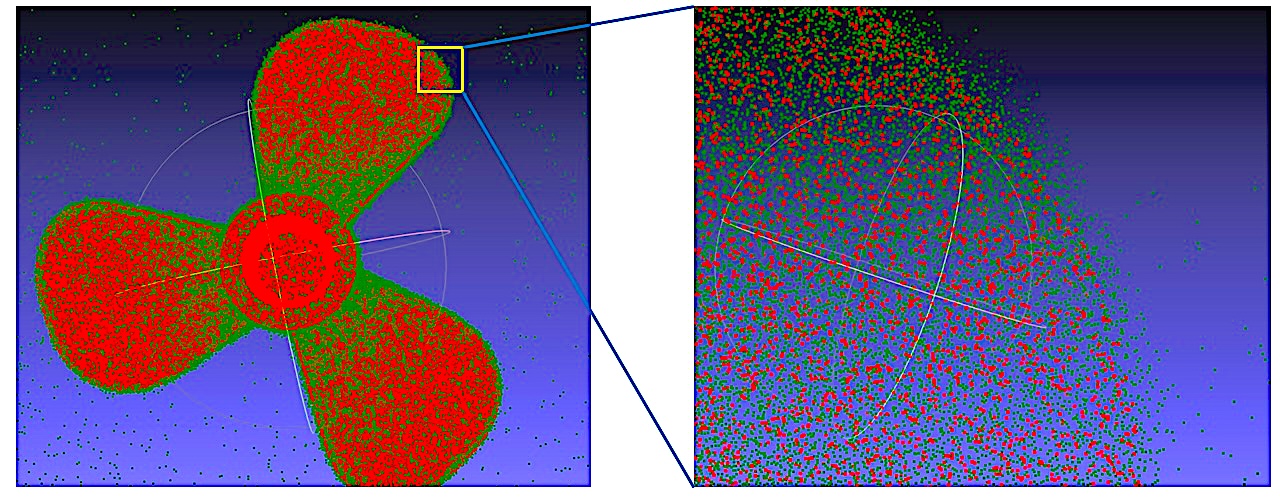}
    \caption{The original point cloud and the point cloud with added noise. The original point cloud is represented in red, while the point cloud with noise is shown in green.}
    \label{fig:cloudwnoise}
\end{figure}

To evaluate the impact of noise on the proposed method, noise was added to the original depth maps incrementally. 
As shown in Table~\ref{table:comp_noise}, $\sigma$ represents the standard deviation of the noise. 
To simulate the characteristic that the measurement error of real depth increases with distance, the standard deviation of the added noise is multiplied by the average depth value of each view.
One can see from the table that as the noise gradually increases, the accuracy of the proposed method gradually diminishes. Interestingly, the method that jointly estimates object pose and point cloud is more susceptible to noise compared to the method only relying on rotational symmetry for object pose estimation. This is primarily due to the reason that increased noise levels weaken the rotational symmetry exhibited by the point clouds, hence leading to a decline in the precision of individual points, though the overall rotational symmetry of the point cloud is still maintained. As shown in Fig.~\ref{fig:cloudwnoise}, under the noise even with $ 2\sigma = 0.002$, a considerable number of points deviated from the original point cloud. It is worth noting that this experiment still demonstrates the capability of the proposed method to leverage rotational symmetry for accurate object pose estimation, even with noise amplified by five times, which verifies its robustness to noise.

\begin{table}[t!]
    \centering
    \caption{Comparison of the ADD-S on the synthetic dataset with different levels of noise. $\sigma = 0.001$. JO denotes joint optimization of object pose and point cloud.}
    \label{table:comp_noise}
{
    \begin{tabular}{l|ccccc}
    \toprule
          Method & $\sigma$ & $2\sigma$ & $3\sigma$ &$4\sigma$ & $5\sigma$  \\ \midrule

        
    \multirow{1}{*}{OURS w/o JO}  & 0.027 & 0.029 & 0.034 & 0.034 & 0.042\\ 
    \multirow{1}{*}{OURS w/ JO} & 0.021 & 0.022 & 0.023 & 0.028 & 0.042\\ 
    \bottomrule
    \end{tabular}
}
\end{table}

\subsubsection{Robustness to occlusion}

\begin{figure}[t!]
    \centering
    \includegraphics[width=\linewidth]{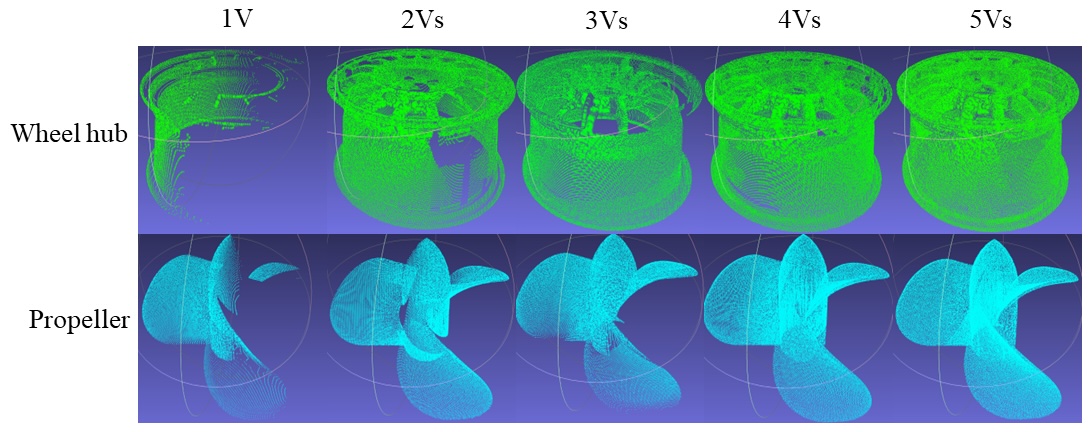}
    \caption{Point clouds with data from varying views. The additional views contribute to a gradual reduction of occlusion-induced missing object structures.}
    \label{fig:viewoculusion}
\end{figure}

In practical scenarios, information loss due to occlusion creates a challenge to pose estimation. Therefore, to evaluate the robustness of the proposed method against occlusion, data from varying views were employed to simulate different degrees of occlusion. As shown in Table~\ref{table:comp_occlusion}, rotational symmetry is maintained even in the presence of occlusion, as long as the majority of the object's surface is observed, as illustrated in Fig.~\ref{fig:viewoculusion}. This substantiates the robustness of the proposed method to occlusion. Moreover, through jointly optimizing point cloud and object pose, the method exhibits good precision in object pose estimation even under occlusion, which further demonstrates the robustness of the proposed approach.

\begin{table}[t!]
    \centering
    \caption{Comparison of ADD-S on the synthetic dataset with data from varying views. JO denotes the joint optimization of the object pose and point cloud.}
    \label{table:comp_occlusion}
{
    \begin{tabular}{l|ccccc}
    \toprule
 Method  & 1V & 2Vs & 3Vs & 4Vs & 5Vs  \\
    \midrule
{OURS  w/o JO}  & 0.078 & 0.032 & 0.051& 0.029& 0.025 \\ 
{OURS w/ JO}    & 0.065 & 0.029 & 0.042& 0.025 & 0.020 \\
    \bottomrule
    \end{tabular}
}
\end{table}

\subsection{Qualitative Analysis}

\begin{figure}[t!]
    \centering
    \includegraphics[width=\linewidth]{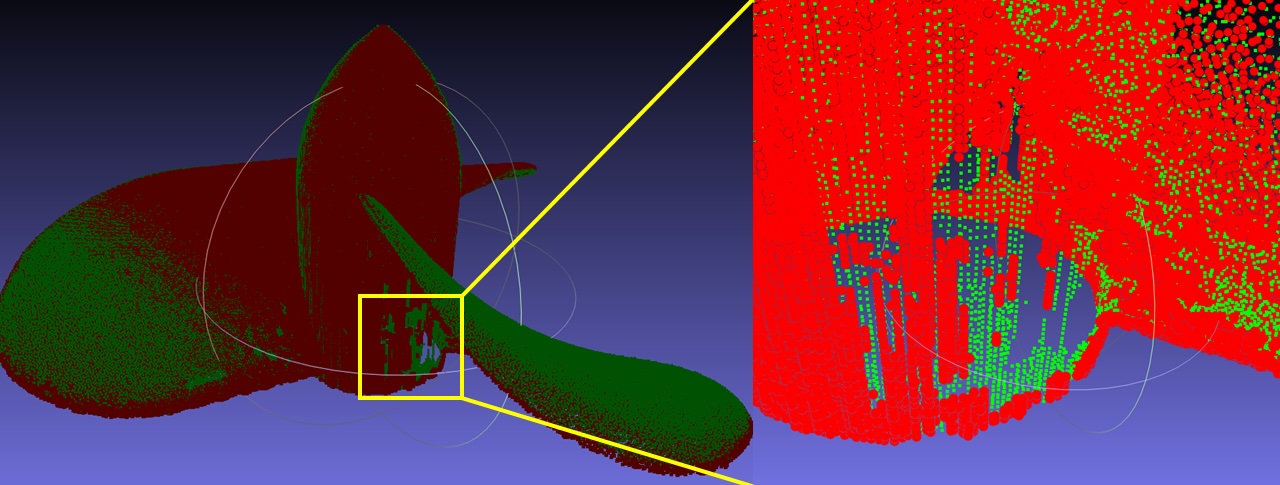}
    \caption{Augmentation of point cloud by rotation completion after the joint optimization. The red points represent the original point cloud, while the green points represent the point cloud patched through rotation completion.}
    \label{fig:argumentedpoints}
\end{figure}

From the results obtained on the synthetic dataset shown in Fig.~\ref{fig:argumentedpoints}, it can be observed that through multiple rotations for completion, the augmented point cloud can utilize information from the unobstructed portion of the point cloud to compensate for the missing element in the occluded parts.

\begin{figure}[t!]
    \centering
    \includegraphics[width=0.8\linewidth]{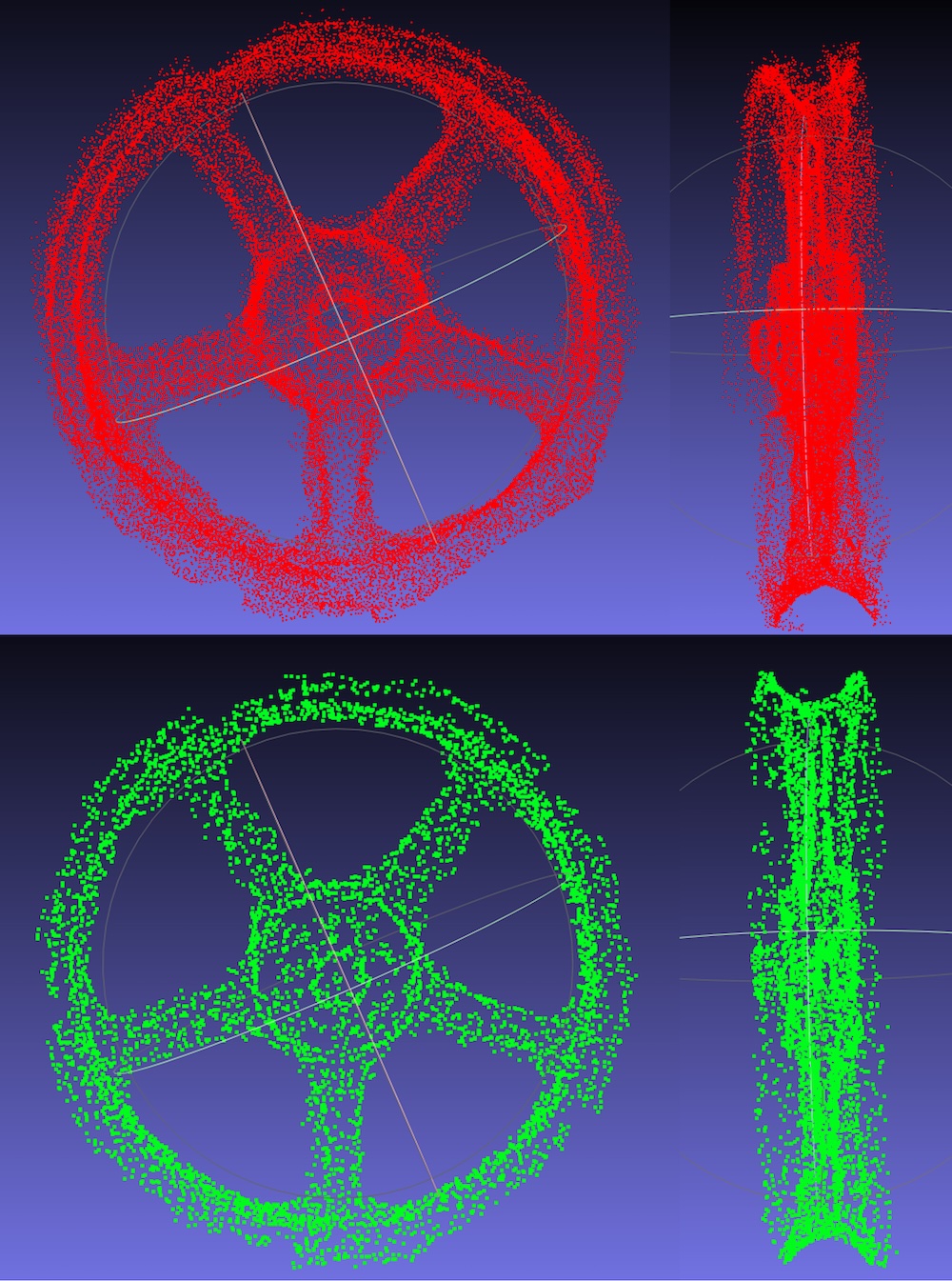}
    \caption{The red dots represent the point cloud after segmentation and outlier removal from the raw point cloud. The green ones represent the point cloud after joint estimation of point cloud and object pose.}
    \label{fig:wheelcomp}
\end{figure}

In addition to the synthetic dataset, the proposed method was also evaluated on an industrial component. The presence of substantial noise and shape distortions in the point cloud presents some challenges. As shown in Fig.~\ref{fig:wheelcomp}, not only did the proposed method accurately estimate the object pose, it also corrected distortions introduced during the SfM point cloud construction.

\subsection{Effectiveness of Joint Optimization}

\begin{figure}[t!]
    \centering
    \includegraphics[width=\linewidth]{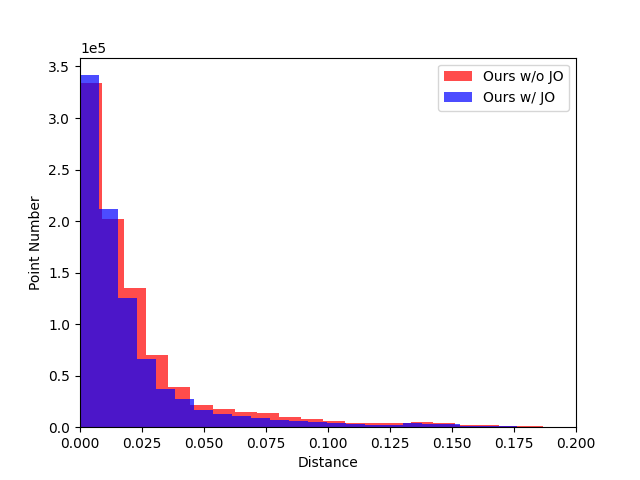}
    \caption{Histogram of the number of points within different error ranges.}
    \label{fig:comp_auc}
\end{figure}

To validate the effectiveness of jointly optimizing pose and point cloud, results from post estimation with and without joint optimization are presented in Fig.~\ref{fig:comp_auc}, which shows that with joint optimization, the number of high-precision points is further increased, thereby improving the accuracy of the final object pose estimation.


\section{Conclusion}
In this paper, we proposed a method for pose estimation from point clouds by utilizing the rotational symmetry of industrial objects. Unlike many existing methods that treat  rotational symmetry as a challenge, the proposed method leverages this property directly. This allows for object pose estimation even in the absence of known 3D object models. Moreover, by optimizing the object point cloud based on multiple observations preserving rotational symmetry, the accuracy of object pose estimation is further enhanced. Finally, by compensating for occluded regions using rotational symmetry, missing areas caused by occlusion can be filled in to obtain more estimation results.

Due to the discrete nature of point cloud models, it is challenging to fully utilize the distribution information inherent in the point cloud during joint optimization. Therefore, future work will focus on utilizing rotational symmetry to further improve the accuracy of three-dimensional reconstruction of objects, particularly in continuous representations such as neural signed distance field.

\bibliography{thebibliography}
\bibliographystyle{IEEEtran}

\newpage

\section{Biography}

\begin{IEEEbiography}[{\includegraphics[width=1in,height=1.25in,clip,keepaspectratio]{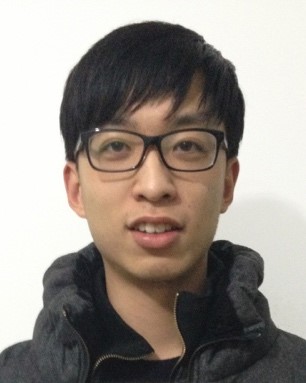}}]
{Weichen Dai} (M'21) was born in Wenzhou, China. He received his B.S. degree in information engineering from Zhejiang University of Technology, Hangzhou, China, in 2015, and his Ph.D. degree from the College of Control Science and Engineering at Zhejiang University, Hangzhou, China, in 2021. His research interests include 3D vision and intelligent autonomous systems. He is now working in the School of Computer Science at Hangzhou Dianzi University, China. 



\end{IEEEbiography}

\begin{IEEEbiography}[{\includegraphics[width=1in,height=1.25in,clip,keepaspectratio]{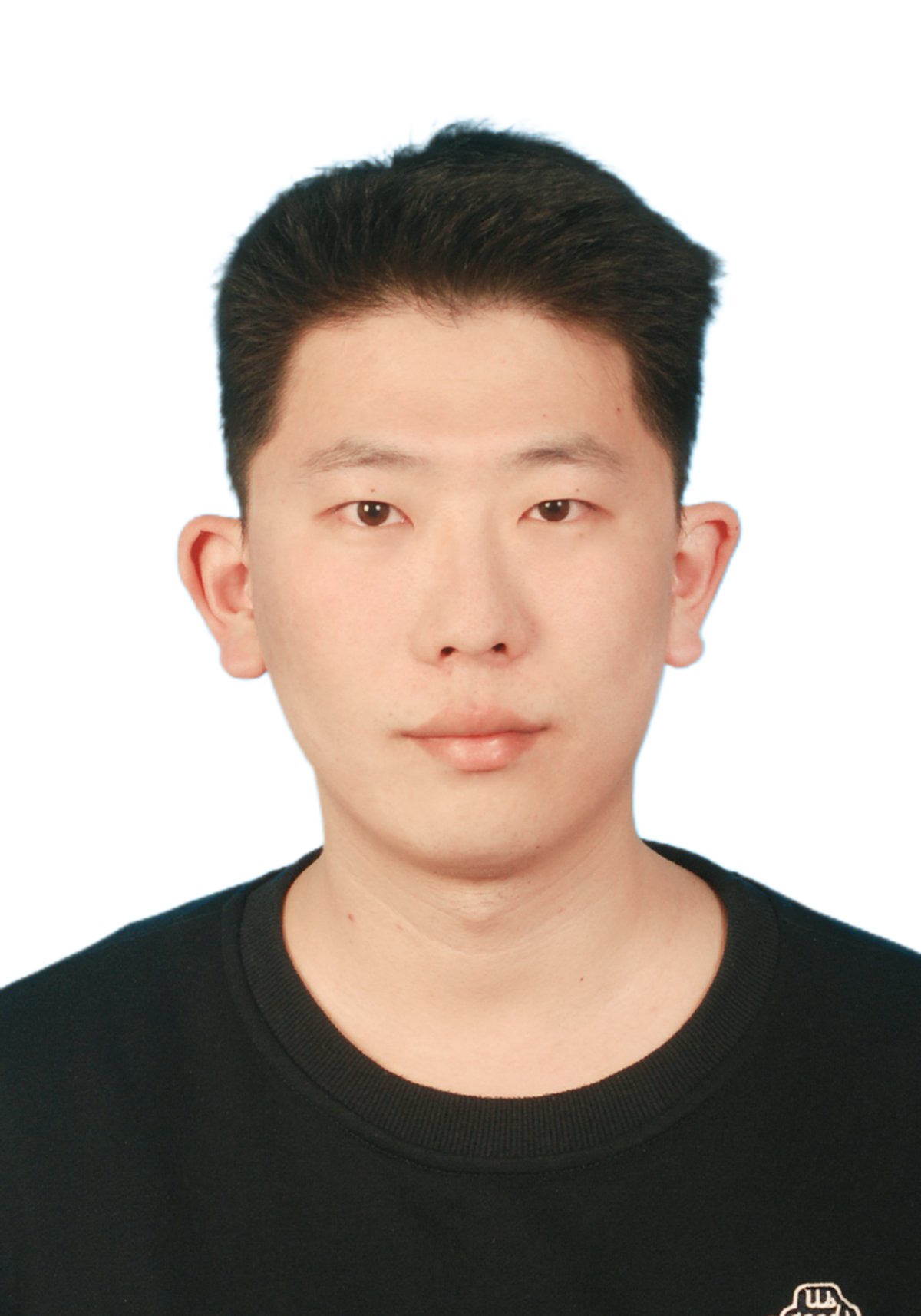}}]
{Ruixun Yu}was born in Lanzhou, China. He received his B.S. degree in computer science and technology from Hangzhou Dianzi University in 2023, and is currently enrolled as a graduate student in the Computer Science program at Hangzhou Dianzi University. His research interests include 3D reconstruction.
\end{IEEEbiography}

\begin{IEEEbiography}
[{\includegraphics[width=1in,height=1.25in,clip,keepaspectratio]{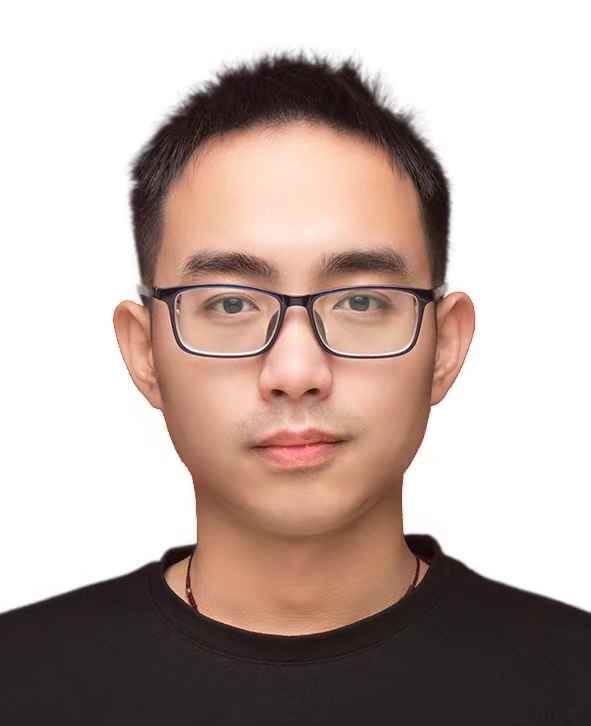}}]
{Yangjie Tang} was born in Huzhou, China. He received his B.Eng. degree in computer science and technology from Zhejiang Agriculture and Forestry University, Hangzhou, China, in 2019, and his M.Eng. degree in computer technology from Hangzhou Dianzi University, Hangzhou, China, in 2024.  His research interests include 3D vision and image processing. 
\end{IEEEbiography}

\begin{IEEEbiography}[{\includegraphics[width=1in,height=1.25in,clip,keepaspectratio]{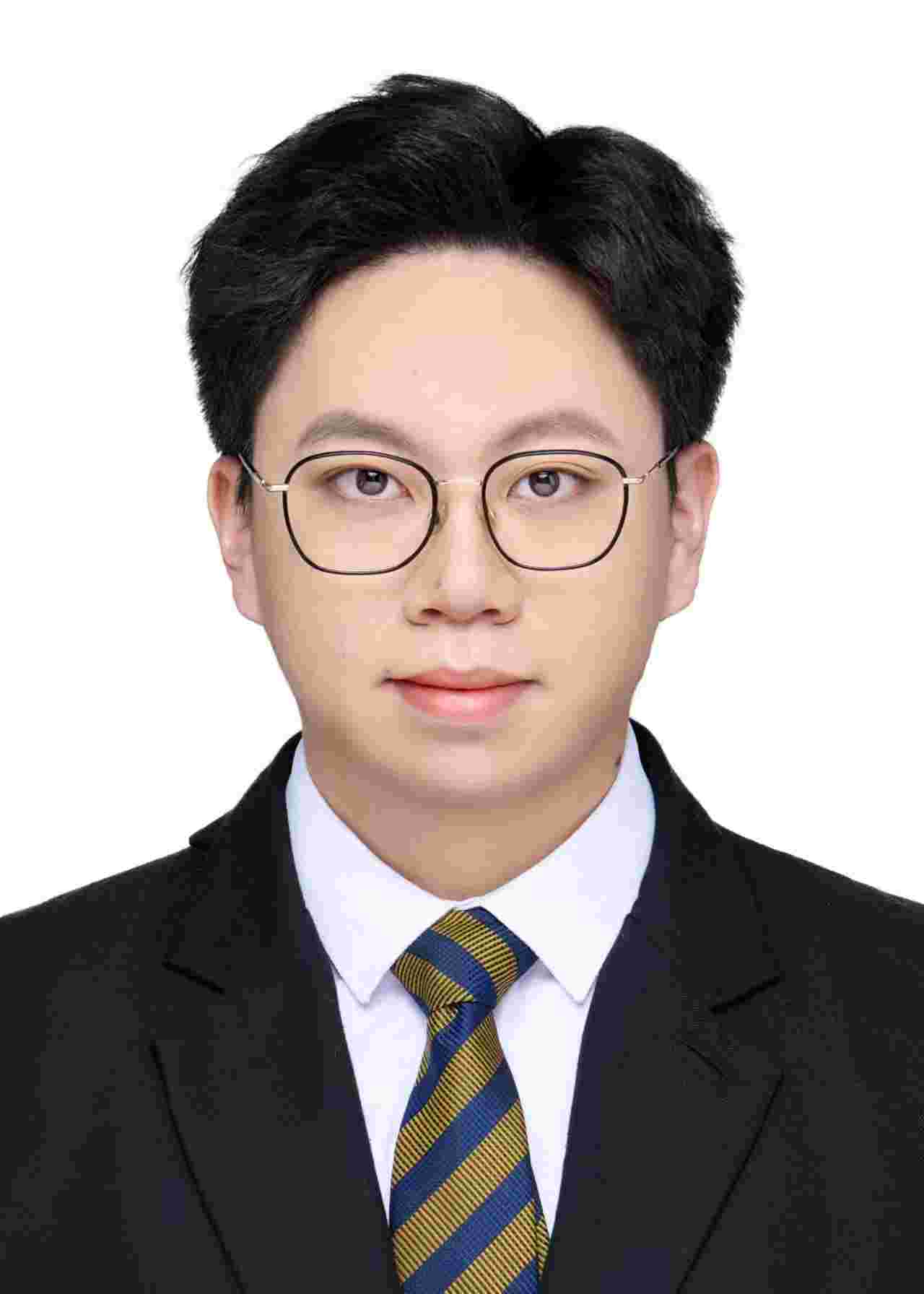}}]
{Yifan Du}was born in JiaXing, China. He received his B.S. degree in computer science and technology from Hangzhou Dianzi University in 2022, and is currently enrolled as a graduate student in the Computer Science program at Hangzhou Dianzi University. His research interest is 3D reconstruction.
\end{IEEEbiography}

\begin{IEEEbiography}[{\includegraphics[width=1in,height=1.25in,clip,keepaspectratio]{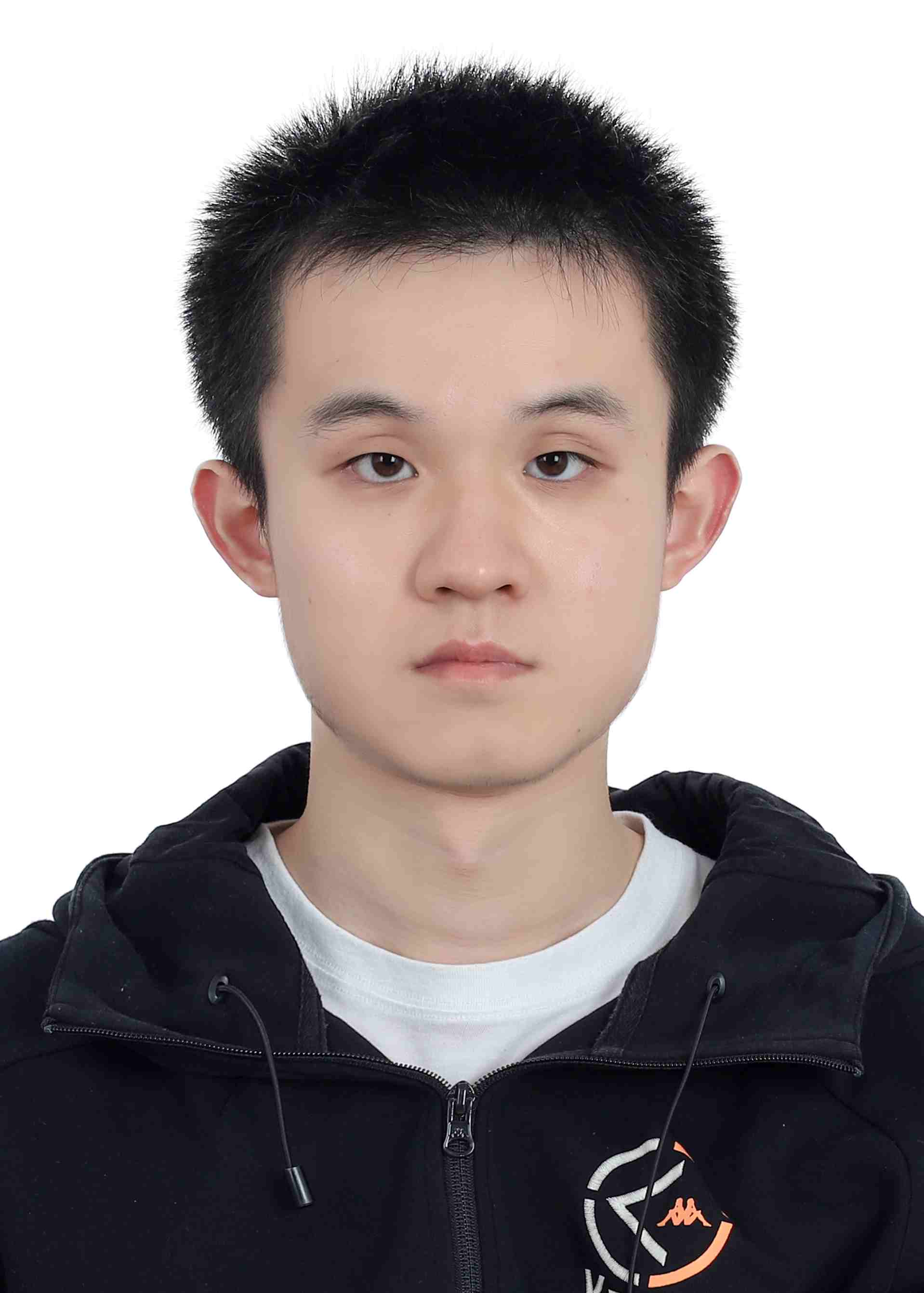}}]
{Yiyang Zhang} was born in Hangzhou, China. He received his B.Eng. degree in computer science and technology from Hangzhou Dianzi University, Hangzhou, China, in 2022. He is currently pursuing his M.Eng degree in computer technology. His research interests include computer vision.
\end{IEEEbiography}

\begin{IEEEbiography}[{\includegraphics[width=1in,clip,keepaspectratio]{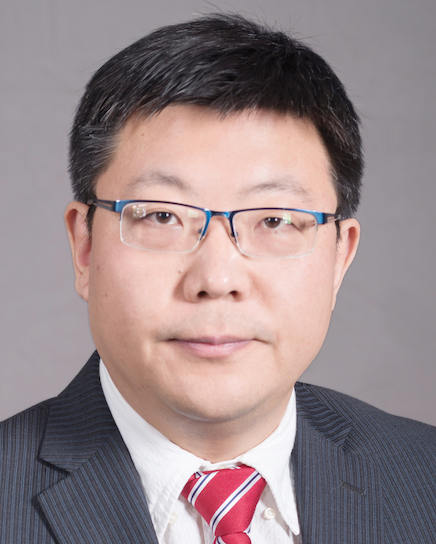}}]{Donglei Sun} (M'15) received the Bachelor's degree in aircraft design and engineering and the Master's degree in vehicle operation engineering from Beihang University, Beijing, China, in 2007 and 2012, respectively, and the Ph.D. degree in aerospace engineering from the University of Illinois at Urbana-Champaign, USA, in 2019. From 2020, he has been with the University of Nottingham Ningbo China, where he is currently Assistant Professor in Aerospace Engineering Design. His research interests include flight control, adaptive control, and aerial robotics.
\end{IEEEbiography}

\begin{IEEEbiography}[{\includegraphics[width=1in,height=1.25in,clip,keepaspectratio]{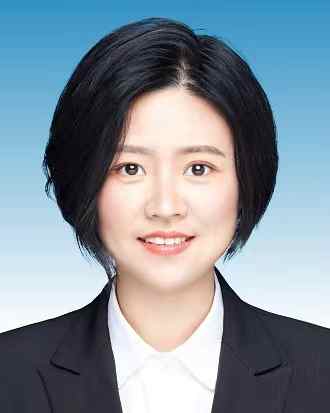}}]
{Hua Zhang} received B.S. and Ph.D. degrees in biomedical engineering and instrumentation from Zhejiang University, China, in 2003 and 2009, respectively. She is currently an associate professor at the School of Computer Science, Hangzhou Dianzi University. Her research interests are image quality assessment, 3D video encoding/decoding, robotic vision and embedded system design.
\end{IEEEbiography}

 




\vfill

\end{document}